\documentclass{article}

\PassOptionsToPackage{numbers, sort}{natbib}
\usepackage[preprint]{neurips_2024}

\usepackage{graphicx}

\usepackage[utf8]{inputenc} 
\usepackage[T1]{fontenc}    
\usepackage{hyperref}       
\usepackage{url}            
\usepackage{booktabs}       
\usepackage{amsfonts}       
\usepackage{nicefrac}       
\usepackage{microtype}      
\usepackage{xcolor}         
\usepackage{amsthm}
\usepackage{amsmath}
\usepackage{amssymb}

\theoremstyle{plain}
\newtheorem{theorem}{Theorem}

\title{Two-Phase Dynamics of Interactions Explains the Starting Point of a DNN Learning Over-Fitted Features}

\author{%
Junpeng Zhang, \\ \textit{Sun Yat-Sen University} \\ zhangjp63@mail2.sysu.edu.cn
\And
Qing Li, \\ \textit{Beijing Institute for General Artificial Intelligence} \\ dylan.liqing@gmail.com
\AND
Liang Lin, \\ \textit{Sun Yat-Sen University} \\ linlng@mail.sysu.edu.cn
\And
Quanshi Zhang\footnote{Quanshi Zhang is the corresponding author. He is with the Department of Computer Science and Engineering,
the John Hopcroft Center, at the Shanghai Jiao Tong University, China.
} \\ \textit{Shanghai Jiao Tong University} \\ zqs1022@sjtu.edu.cn
}

\begin{document}

\maketitle

\begin{abstract}
    This paper investigates the dynamics of a deep neural network (DNN) learning interactions. Previous studies have discovered~\cite{li2023does} and mathematically proven~\cite{ren2023we} that given each input sample, a well-trained DNN usually only encodes a small number of interactions (non-linear relationships) between input variables in the sample. A series of theorems have been derived to prove that we can consider the DNN's inference equivalent to 
    using these interactions as primitive patterns for inference.
    In this paper, we discover the DNN learns interactions in two phases.
    The first phase mainly penalizes interactions of medium and high orders, and the second phase mainly learns interactions of gradually increasing orders. We can consider the two-phase phenomenon as the starting point of a DNN learning over-fitted features. Such a phenomenon has been widely shared by DNNs with various architectures trained for different tasks. 
    Therefore, the discovery of the two-phase dynamics provides a detailed mechanism for how a DNN gradually learns different inference patterns (interactions). In particular, we have also verified the claim that high-order interactions have weaker generalization power than low-order interactions. Thus, the discovered two-phase dynamics also explains how the generalization power of a DNN changes during the training process.
\end{abstract}

\section{Introduction}\label{sec:intro}
Most existing studies~\cite{dziugaite2017computing, neyshabur2015norm, keskar2016large, foret2020sharpness, kwon2021asam} considered the generalization power of an AI model as an intrinsic property of the \textbf{entire} model. However, in this study, let us revisit the generalization power of DNNs from a new perspective, \emph{i.e.,} directly quantifying the primitive inference patterns encoded by a DNN to explain the DNN's overfitting. There have been two substantial progresses in this new direction in recent years.

\begin{figure}[t]
    \centering
    \includegraphics[width=\linewidth, height = 6cm]{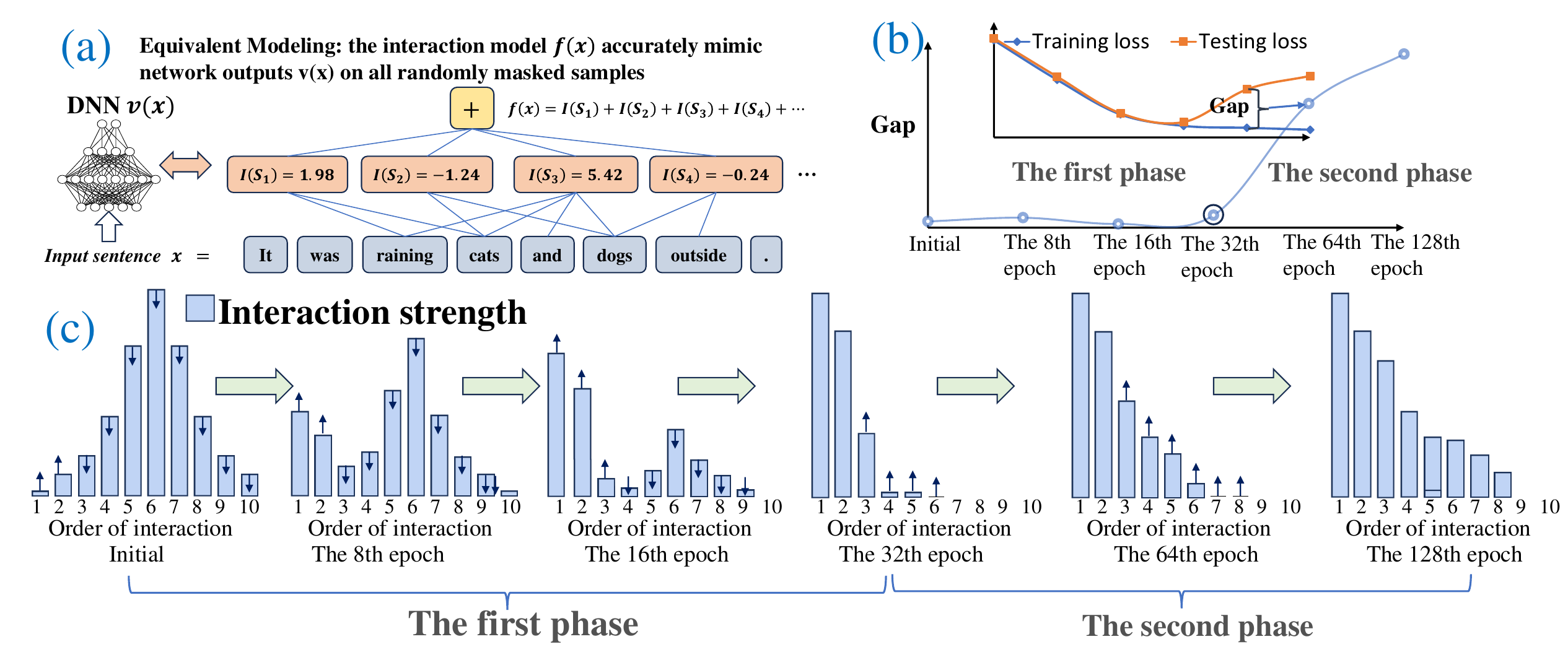}
    \caption{(a) Illustration of interactions encoded by a DNN. Each interaction is a metric to measure a non-linear relationship among a specific set $S$ of input variables. (b) The two-phase dynamics is temporally aligned with the change of the gap between the testing and training losses during the learning process. (c) Illustration of the two-phase phenomenon.}
    \label{img-fig1}
\end{figure}

\textbullet \ \textbf{Background 1: Proving that the complex inference score of a DNN can be faithfully explained by \textit{primitive inference patterns}.}
Explaining the inference of a DNN as symbolic inference patterns is a fundamental yet counter-intuitive problem in the field of explainable artificial intelligence (XAI). Fortunately, it is experimentally discovered~\cite{li2023does} and mathematically proved~\cite{ren2023we} that given a specific input sample, a well-trained DNN usually only encodes a small number of interactions. Each interaction is a metric to measure a non-linear relationship between a specific set of input variables in $S$. As Figure \ref{img-fig1} shows, the DNN may encode an interaction between four words of $S = \{\text{raining}, \text{cats}, \text{and}, \text{dogs}\}$  (\emph{i.e.,} four input variables) in the input sentence. The co-appearance of the four words triggers this interaction and makes an effect $I(S)$ that pushes the DNN's output towards the meaning of ``heavy rain.'' \textbf{It is proven~\cite{ren2023defining, ren2023we, li2023does} that almost all subtle changes of network outputs w.r.t. any random masking of input variables can all be mimicked by these interaction effects just like primitive inference patterns.}

\textbullet \ \textbf{Background 2: The interaction enables us to analyze the specific generalization power of each input sample in a much more fine-grained manner,} instead of simply projecting the entire input sample into a single high-dimensional feature. To this end, \citet{zhou2024explaining} have found that high-order interactions often have weaker generalization power than low-order interactions. The order of an interaction is defined as the number of variables in the interaction, order$(S) =\vert S \vert$,
to represent the complexity of the interaction. 

\textbf{Discovering the two-phase phenomenon of the change of interaction’s complexity during the training process.}
Based on the above findings, in this paper, we hope to explore a new issue, \emph{i.e.,} identifying the exact starting point (epoch) when this epoch begins to learn over-fitted features. Specifically, as figure \ref{img-fig1}, we discover a two-phase phenomenon in the training process of DNNs, which reveals the hidden factors that push the DNN from under-fitting to over-fitting throughout the entire training process.

\textit{(1) Before the training process:} a DNN with initialized parameters mainly encodes interactions of medium orders and seldom encodes interactions of very high orders or very low orders. The distribution of interactions over different orders looks like a fusiform.

\textit{(2) In the first phase:} higher-order interactions are gradually eliminated. Eventually, at the end of the first phase, the DNN mainly encodes only low-order interactions.

\textit{(3) In the second phase:} the DNN learns interactions of gradually increasing orders. The gradual increase of the interaction order is a typical phenomenon of learning over-fitted features.

\textbf{Alignment between the two-phase training process and the loss gap.} When we use the gap between training and testing losses to measure the overfitting level, we find that the dynamics of the overfitting level is consistently aligned with the dynamics of the two-phase training process. As Figure \ref{img-fig1} shows, during the entire first phase, the gap between the training loss and the testing loss is relatively small, \emph{i.e.,} the DNN has not learned over-fitted features. Whereas, shortly after entering the second phase, the gap widens rapidly, which indicates the beginning of learning over-fitted features.
 
\textbf{Identifying the starting point of learning over-fitted features.} 
Unlike traditionally considering overfitting as a property of the entire model, we aim to identify the starting point of overfitting for each specific training sample. \textbf{Typically, the starting point of learning over-fitted features for the DNN occurs shortly into the second phase.} Note that the starting point does not mean the entire model has been significantly biased. Instead, both the generalizable features and non-generalizable (over-fitted) features are learned simultaneously in the second phase, and the testing accuracy still keeps increasing.
Specifically, in the second phase, the DNN begins to shift its attention from exclusively learning low-order interactions to learning interactions of bit higher orders. Therefore, we consider the DNN actually learns more complex interactions from under-fitting towards over-fitting in the second phase.

More crucially, the above two-phase phenomenon is widely observed in various DNNs trained for different tasks. To this end, we conducted experiments on seven different DNNs trained on six different datasets, towards image classification, natural language processing, and 3D point cloud classification. In these experiments, we both verified the relationship between the order of the interactions and verified their generalization power and the two-phase dynamics of learning interactions.

In sum, although previous work~\cite{chen2024defining, liu2024towards, zhou2024explaining} has found that the inference score of a DNN can be explained as the sum of the effects of a small number of interactions, and has found the weak generalization power of high-order interactions, our study has substantially extended the understanding of the generalization power of DNNs.
The discovered two-phase dynamics of interactions explains the detailed mechanical factors that potentially determine how the generalization power changes during the training process.
Specifically, in the first phase, the DNN removes the noise and learns low-order interactions to improve generalization power. In the second phase, the DNN gradually learns more complex interactions, and feature representations of the DNN gradually change from underfitting to overfitting. In addition, our research also shows that the two-phase phenomenon is temporally aligned with the dynamics of the gap between the testing and training losses. We believe that the above findings explain the starting point of a DNN learning over-fitted features. 

\section{Related work}\label{sec:relate}
Post-hoc explanation of DNNs is a classical direction of explainable AI. However, the disappointing view of the faithfulness of post-hoc explanation of a DNN has been widely accepted for years~\cite{rudin2019stop, adebayo2018sanity, ghassemi2021false}. Fortunately, recent progress~\cite{ren2023defining, ren2023we, li2023does} in explaining interactions encoded by a DNN provides a new perspective to analyze the DNN, and a series of properties (\emph{e.g.,} sparsity property and universal matching property) of interactions are proven to mathematically guarantee that the interactions can faithfully represent primitive inference patterns of the DNN.

The theoretical system of interactions mainly makes breakthroughs in the following three aspects. (1) It has been mathematically proven~\cite{ren2023we} that a DNN’s inference logits can be faithfully explained through interactions, \emph{i.e.,} given a specific input sample, a sufficiently-trained DNN usually only encodes a small number of interactions for inference, and the DNN's outputs on all different masked states of the sample can be universally matched by these interactions. (2) It has been demonstrated that interactions can explain the generalization power~\cite{zhou2024explaining, zhang2021interpreting}, and the adversarial robustness/transferability of a DNN~\cite{wang2021interpreting, ren2023bayesian}. (3) It has been found that twelve approaches for improving adversarial transferability all share a common mechanism~\cite{ren2021unified}, i.e., they are implicitly reducing interactions between adversarial perturbations, while fourteen attribution methods can all be explained as re-allocation of interaction effects~\cite{wang2020unified}.

Therefore, in this study, we further develop the interaction theory to directly use the primitive inference patterns encoded by a DNN to explain the generalization power of the DNN,  \emph{i.e.,} identifying the exact starting point of learning over-fitted features from each input sample. Compared to traditional analysis of the generalization power of a DNN, we believe that decomposing a DNN's exact inference score into interactions provides much deeper insights into the essential factors that determine a DNN’s generalization power. 
First, the previous work~\cite{dziugaite2017computing, keskar2016large} mainly analyzed already converged DNNs, instead of estimating the time of starting learning over-fitted features. Therefore, we explore the dynamics of the order (complexity) of interactions encoded by a DNN to analyze the change of the generalization power of a DNN.
Second, previous work considered overfitting as an overall property of a model. For example, explaining DNNs through the theoretical bounds for the generalization~\cite{dziugaite2017computing, neyshabur2015norm}, smoothness of the loss landscape~\cite{keskar2016large, foret2020sharpness, kwon2021asam} often project each sample to a high-dimensional feature point in the feature space. However, we find that 
such a claim is usually valid for shallow models, but for complex DNNs, different training samples usually have different starting points of overfitting. Thus, we propose to evaluate whether a DNN has begun to learn over-fitted features by evaluating the distribution of interactions over different orders.

\section{Analyzing the starting point of learning over-fitted features}
\subsection{Preliminaries: interactions}
Let us consider a DNN $v$ and an input sample $\mathbf{x} = [x_1, x_2, ..., x_n]^T$, which contains $n$ input variables and is indexed by $N = \{1, 2, ..., n\}$. Let $v(\mathbf{x})\in \mathbb{R}$ be a scalar\footnote{The DNN's output score can be defined in different forms. For example, for multi-class classification, $v(\mathbf{x})$ can be defined as either $v(\mathbf{x}) = \log \frac{p(y=y^{*}|\mathbf{x})}{1-p(y=y^{*}|\mathbf{x})}$ or the scalar output corresponding to the ground-truth label before the softmax layer, where $p(y=y^{*}|\mathbf{x})$ represents the probability of the ground-truth label.} output of the DNN. We can decompose the $v(\mathbf{x})$ into a set of AND interactions $I_{\text{\text{and}}}(\mathbf{x}) \in \Omega_{\text{and}}$ and a set of OR interactions $I_{\text{or}}(\mathbf{x}) \in \Omega_{\text{or}}$ by following \citet{zhou2023explaining}, as follows,
\begin{align}\label{equation:1}
    v(\mathbf{x}) = \sum\nolimits_{S \in \Omega_{\text{and}}} I_{\text{and}}(S\vert \mathbf{x}) + \sum\nolimits_{S \in \Omega_{\text{or}}}I_{\text{or}}(S\vert \mathbf{x}) + v(\emptyset).
\end{align}

The above AND-OR interactions can be understood as follows. Theorem \ref{theorem:2} shows that an AND interaction represents an AND relationship between input variables in $S \subseteq N$ encoded by the DNN (\emph{e.g.,} the compositional relationship between image regions encoded for image classification or the interactions between words for natural language processing). For example, given an input sentence $\mathbf{x} = \text{``it was raining cats and dogs outside,''}$ the DNN may encode an interaction between $S=\{ \text{raining}, \text{cats}, \text{and}, \text{dogs}\}$ to make an effect $I(S \vert \mathbf{x})$ that pushes the DNN's output towards the meaning of ``heavy rain.'' If any input variable in $S$ is masked\footnote{\label{footnote: baseline}A variable in $S$ is masked means that the variable in $S$ is replaced by a baseline value. The baseline value is usually set to the mean value of this variable over different samples~\cite{dabkowski2017real}.}, this numerical effect will be removed from the DNN's output. 
Similarly, an OR interaction $I_{\text{or}}(S\vert \mathbf{x})$ represents an OR relationship between input variables in $S$. For example, given an input sentence $\mathbf{x} = \text{``the movie is disappointing and uninspiring,''}$ the presence of either word in $S = \{\text{disappointing}, \text{uninspiring}\}$  will push the DNN's output towards the meaning of negative sentiment. The AND interaction and the OR interaction for $\forall \emptyset \not= S\subseteq N$ are defined as
\begin{align}
    I_{\text{and}}(S \vert \mathbf{x}) \overset{\triangle}{=} \sum\nolimits_{T \subseteq S} (-1)^{\vert S \vert - \vert T \vert} v_{\text{and}}(\mathbf{x}_T), \ 
    I_{\text{or}}(S \vert \mathbf{x}) \overset{\triangle}{=} - \ \sum\nolimits_{T \subseteq S} (-1)^{\vert S \vert - \vert T \vert} v_{\text{or}}(\mathbf{x}_{N \backslash T}),
\end{align}
where $\mathbf{x}_T$ represents a masked input sample in which we mask\textsuperscript{\ref{footnote: baseline}} input variables in $N \setminus T$. The DNN's output $v(\mathbf{x}_T)$ is decomposed into two components $v(\mathbf{x}_T) = v_{\text{and}}(\mathbf{x}_T) + v_{\text{or}}(\mathbf{x}_T) + v(\emptyset)$. It is easy to prove\footnote{See appendix \ref{sec:match} for the detailed proof.} that the component $v_{\text{and}}(\mathbf{x}_T)$ only contains AND interactions $v_{\text{and}}(\mathbf{x}_T) = \sum_{\emptyset \ne S^{\prime} \subseteq T} I_{\text{and}}(S^{\prime}\vert \mathbf{x}_T)$, and the component $v_{\text{or}}(\mathbf{x}_T)$ only contains OR interactions $v_{\text{or}}(\mathbf{x}_T) = \sum_{S\subseteq N:S \cap T \neq \emptyset} I_{\text{or}}(S\vert \mathbf{x}_T)$.

\textbf{How to compute AND-OR interactions.} 
In order to compute AND-OR interactions, we set the component $v_{\text{and}}(\mathbf{x}_T) = 0.5\cdot v(\mathbf{x}_T) - 0.5\cdot v(\emptyset) + \gamma_T$ and set $v_{\text{or}}(\mathbf{x}_T) = 0.5\cdot v(\mathbf{x}_T) - 0.5\cdot v(\emptyset) - \gamma_T$. The parameters $\{\gamma_T\}$ are learned to obtain the sparsest interactions via $\mathop{\min}_{\{\gamma_T\}} \sum_{S} \vert I_{\text{and}}(S\vert \mathbf{x}) \vert + \vert  I_{\text{or}}(S\vert \mathbf{x}) \vert$~\cite{li2023defining}.

\begin{figure}[t]
    \centering
    \includegraphics[width=0.99\textwidth]{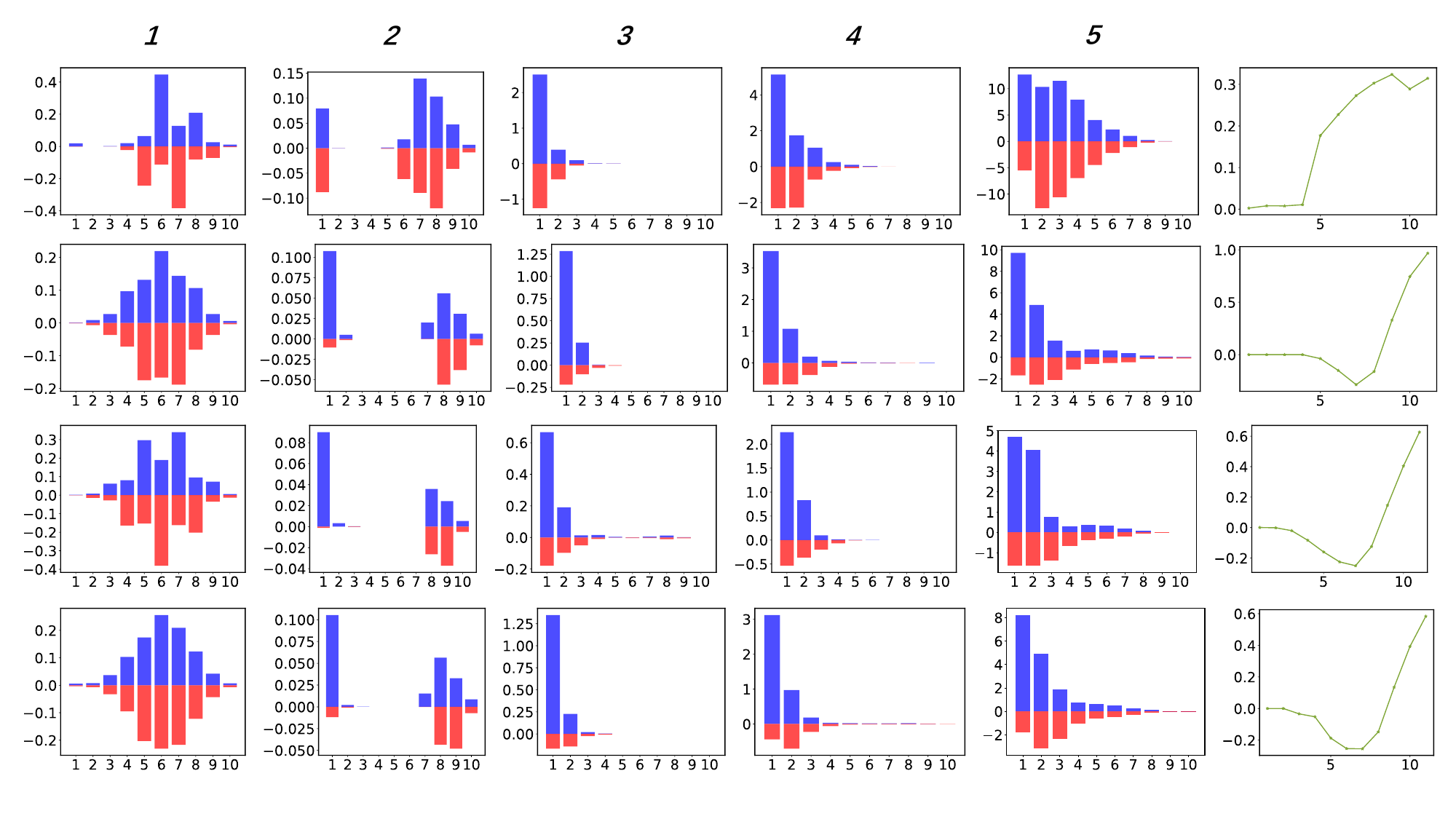}
    \caption{The positive and negative strength of salient interactions over different orders. Experiments show that various DNNs trained on different datasets for different tasks all exhibited the two-phase phenomenon, and the two-phase phenomenon is temporally aligned with the change of the gap between the testing and training losses.}
    \label{img-two-phase-experiment}
\end{figure}

\textbf{Theoretical guarantee of taking interactions as a faithful explanation of the DNN's inference logic.} 
The following four properties ensure the faithfulness of interaction-based explanation.

\textbullet \ \textit{Sparsity property.} Theorem \ref{theorem:1} shows that 
the inference of a sufficiently trained DNN is equivalent to an interaction-based model using a small number of interactions for inference.
 Theoretically, there are totally $2^n$ different subsets $S\subseteq N$ of input variables, but only a few subsets of input variables have salient interaction effects. The DNN only encodes a few salient AND interactions in $\Omega_{\text{and}} = \{S\subseteq N :\vert I_{\text{and}}(S \vert x)\vert >\tau \}$ and a few salient OR interactions in $\Omega_{\text{or}} = \{S\subseteq N :\vert I_{\text{or}}(S \vert x)\vert >\tau \}$, where $\tau$ represents a threshold. More precisely, there are only $\mathcal{O}(n^{\kappa} / \tau) \ll 2^n$ interactions with absolute effects greater than the threshold $\tau$. These interactions are sparse, because $\kappa$ empirically ranges in $[0.9, 1.2]$.

\textbullet \ \textit{Universal matching property.} Theorem \ref{theorem:2} shows that given a specific input sample $\mathbf{x}$, there are a total of $2^n$ binary masks corresponding to $2^n$ different subsets $S \subseteq N$ of input variables. Then, the DNN's outputs $v(\mathbf{x}_S)$ on all masked samples can be accurately matched by the numerical effects of a small number of salient interactions.


\textbullet \ \textit{Transferability property.}
Empirical studies~\cite{li2023does} have demonstrated the transferability of interactions across different input samples in classification tasks. Specifically, it has been observed that there are a set of common salient interactions shared by different samples in the same category, \emph{i.e.,} these interactions are frequently extracted by the DNN from different input samples.

\textbullet \ \textit{Discrimination property.} Empirical studies~\cite{li2023does} also have demonstrated the discrimination property of interactions in classification tasks. As mentioned above, a specific salient interaction $I(S\vert \mathbf{x})$ can be extracted from different samples. Then, this interaction usually pushes the classification score to the same category, i.e., the numerical effects of this interaction on different samples are usually consistently positive (or consistently negative) to the classification score.
\begin{theorem}[Sparsity property, proved by \citet{ren2023we}] \label{theorem:1}
Let us be given an input sample $\mathbf{x}$ with $n$ input variables. Let us use a threshold $\tau$ to select a set of salient AND interactions $\Gamma$, subject to $\vert I_{\text{and}}(S\vert \mathbf{x})\vert > \tau$. If the DNN's outputs score $v(\mathbf{x}_{T})$ on all $2^n$ samples $\{\mathbf{x}_T \vert T \subseteq N\}$ are relatively stable\footnote{
The relatively stable output of a DNN on input samples with different masking $\{\mathbf{x}_T \vert T \subseteq N\}$ can be represented as the following three conditions in mathematics. (1) High-order interactions were not encoded by the DNN. (2) Let $\{\mathbf{x}_T:\vert T \vert=n-m\}$ denotes the set of masking samples, where we mask $m$ input variables. In this way, the average output score of the DNN  $\mathbb{E}_{T}[v(\mathbf{x}_T)]$ over $\{\mathbf{x}_T:\vert T \vert=n-m\}$ monotonically decreases as $m$ increases. (3) The decreasing speed of $\mathbb{E}_{T}[v(\mathbf{x}_T)]$ is polynomial. Please see Appendix \ref{sec:condition} for details.}, then the upper bound  of the number of salient interactions $\vert \Gamma \vert$ is $\mathcal{O}(\frac{n^{\kappa}}{\tau})$.
\end{theorem}

\begin{theorem}[Universal matching, proved by ~\citet{zhou2023explaining}]\label{theorem:2}
A DNN’s outputs on all $2^n$ masked samples $\{x_T \vert T \subseteq N\}$ can be universally matched by the numerical effects of a small number of salient interactions.
\begin{align}
    v(\mathbf{x}_T) &= \underbrace{\sum\nolimits_{S\subseteq N} I_{\text{and}}(S \vert \mathbf{x}) \cdot \mathbf{1}(\substack{\mathbf{x}_{\text{T}}\  {\scriptstyle\text{triggers AND}} \\ {\scriptstyle\text{relationship}} \ S})}_{v_{\text{and}}(\mathbf{x}_T)} +  \underbrace{\sum\nolimits_{S \subseteq N} I_{or}(S\vert \mathbf{x}) \cdot \mathbf{1}(\substack{\mathbf{x}_{\text{T}}\  {\scriptstyle\text{triggers OR}} \\ {\scriptstyle\text{relationship}} \ S})}_{v_{\text{or}}(\mathbf{x}_T)} + v(\mathbf{x}_\emptyset)
    \\ &= \sum\nolimits_{\emptyset \ne S \subseteq T}I_{\text{and}}(S\vert \mathbf{x}) + \sum\nolimits_{S \subseteq N: S\cap T \ne \emptyset} I_{or}(S\vert \mathbf{x}) + v(\mathbf{x}_\emptyset) \\
    &\approx  \sum\nolimits_{S \in \Omega_{\text{and}}:S \subseteq T}I_{\text{and}}(S\vert \mathbf{x}) + \sum\nolimits_{S \in \Omega_{\text{or}}:S \cap T \ne \emptyset} I_{or}(S \vert \mathbf{x}) + v(\mathbf{x}_\emptyset).
\end{align}
\end{theorem}

\textbf{The order of an interaction.} We define the order of an interaction $I(S \vert \mathbf{x})$ as the number of input variables in $S \subseteq N$, \emph{i.e.,} $\text{order}(S) = \vert S \vert$, which reflects the complexity of the interaction. 
Specifically, the lower order means that the interaction contains fewer input variables, thereby being simpler.


\subsection{The two-phase phenomenon}
In this study, we analyze the dynamics of the generalization power of a DNN during the training process. Specifically, since \citet{zhou2024explaining} have found that high-order interactions have weaker generalization power than low-order interactions, we aim to analyze the change of the distribution of salient interactions over different orders during the training process, in order to analyze the change of the generalization power of the DNN.

As Figure \ref{img-two-phase-experiment} shows, the distribution of interactions over different orders is quantified by the strength of interactions over different orders. Specifically, due to the sparsity property of interactions, we only focus on salient interactions, which are considered primitive inference patterns encoded by the DNN, and ignore non-salient ones. Salient interactions are defined as $\Omega_{\text{and}}=\{S\subseteq N:\vert I_{\text{and}}(S \vert \mathbf{x})\vert >\tau \}$ and $\Omega_{\text{or}} = \{S\subseteq N :\vert I_{\text{or}}(S \vert \mathbf{x})\vert >\tau \}$, where $\tau$ represents a threshold. Then, given all salient AND-OR interactions of the $k$-th order, we quantify the strength of all positive salient interactions $J_{\text{pos}}^{(k)}$ and the strength of all negative salient interactions $J_{\text{neg}}^{(k)}$, as $J_{\text{pos}}^{(k)} = \sum\nolimits_{S \in \Omega_{\text{and}} \cup \Omega_{\text{or}}:\vert S \vert = k} \max(I(S \vert \mathbf{x}), 0)$ and $J_{\text{neg}}^{(k)} = -\sum\nolimits_{S \in \Omega_{\text{and}} \cup \Omega_{\text{or}}: \vert S \vert = k} \vert \min(I(S \vert \mathbf{x}), 0) \vert$. In this way, Figure \ref{img-two-phase-experiment} visualizes the strength of interactions over different orders encoded by DNNs trained at different epochs, which reflect the following two-phase dynamics during the training process.

\textbullet \ \textit{Before the training process, } as Figure \ref{img-two-phase-experiment} shows, 
an initialized DNN encodes mostly medium-order interactions and rarely encodes high-order and low-order interactions, and the distribution of interactions over different orders looks like a fusiform.
Considering that the DNN with initialized parameters encodes noisy patterns, we can prove the fusiform-like distribution of interactions over different orders. These interactions are extracted from output noises of the initialized DNN, thereby having little generalization power.

\textbullet \ \textit{In the first phase of the training process,} 
as Figure \ref{img-two-phase-experiment} shows, the strength of the high-order and medium-order interactions encoded by the DNN gradually decreases, while the strength of the low-order interactions gradually increases. Eventually, the high-order and medium-order interactions are gradually eliminated, and the DNN encodes only low-order interactions. 
In comparison, the low-order interactions learned during the first phase usually have high generalization power. Therefore, the first phase can be considered mainly to eliminate noisy high-order and medium-order interactions, and gradually the DNN only encodes the simplest low-order interactions with high generalization power.

\textbullet \ \textit{In the second phase of the training process,} as Figure \ref{img-two-phase-experiment} shows, the DNN encodes interactions of gradually increasing orders (complexity) during the training process. Gradually learning more and more complex interactions makes the over-fitting risk gradually increase. Thus, this is a slow process from under-fitting to over-fitting.

\begin{figure}[t]
    \centering
    \includegraphics[width=0.99\textwidth]{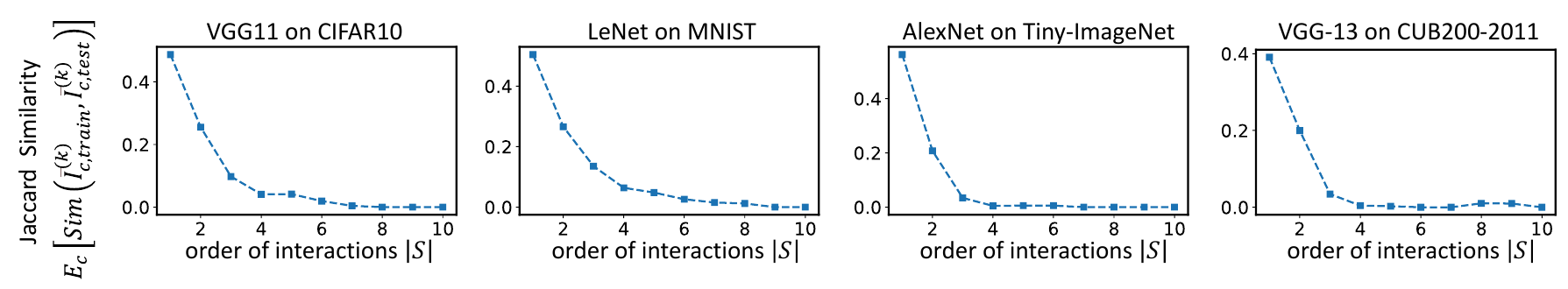}
    \caption{The mean of the Jaccard similarity over the first $10$ categories between interactions extracted from training samples and that extracted from testing samples.}
    \label{img-similarity}
\end{figure}

\subsection{Experiments}\label{sec:experiments}
\textbf{\textbullet \ Is the two-stage phenomenon during the training process ubiquitously observed on most DNNs on different datasets for various tasks? 
} 
We trained LeNet~\cite{lecun1998gradient}, VGG-11/13/16~\cite{simonyan2014very}, and AlexNet~\cite{krizhevsky2012imagenet} on image datasets (the CIFAR-10 dataset~\cite{krizhevsky2009learning}, the MNIST dataset~\cite{lecun1998gradient}, the CUB200-2011 dataset~\cite{wah2011caltech} with cropping background regions around the bird, and the Tiny-ImageNet dataset~\cite{tiny-imagenet}), while we trained the Bert-Medium/Tiny model~\cite{devlin2018bert} on natural language dataset (the SST-2 dataset~\cite{socher2013recursive}), and we trained DGCNN~\cite{dgcnn} on 3D point cloud dataset (the ShapeNet dataset~\cite{chang2015shapenet}). Figure \ref{img-two-phase-experiment} shows the distribution of salient interactions over different orders extracted from different DNNs at different training epochs. We found the two-stage phenomenon on all these DNNs.

\textbf{\textbullet \  Do high-order interactions have weaker generalization power than low-order interactions?} If an interaction frequently extracted from the training samples can also be frequently observed in testing samples, then this interaction is considered generalizable to testing samples. Thus, we follow \citet{zhou2024explaining} to use the Jaccard similarity between the distribution of interactions extracted from training samples and the distribution of interactions extracted from testing samples to measure the generalization power of interactions. Specifically, 
given each input sample $\mathbf{x}$ with $n$ input variables, we vectorize all interactions of $k$-th order extracted from $\mathbf{x}$ as $\mathbf{I}^{(k)}(\mathbf{x})=[I(S_{1} \vert \mathbf{x}), I(S_{2} \vert \mathbf{x}), ..., I(S_{d} \vert \mathbf{x})]^T$, where $S_1, S_2, ..., S_d$ denotes the all $d = \tbinom{n}{k}$ interactions of the $k$-th order. Then, we compute the average interaction vector of $k$-th order over all samples in the category $c$ as $\overline{\mathbf{I}}^{(k)}_{c}=\mathbb{E}_{\mathbf{x} \in C}[\mathbf{I}^{(k)}(\mathbf{x})]$ to represent the distribution of interactions of $k$-th order extracted from samples in the category $c$, where $C$ denotes the set of samples in the category $c$. Then, we compute the Jaccard similarity between the average interaction vector of $k$-th order over training samples $\overline{\mathbf{I}}^{(k)}_{c, \text{train}}$ and the average interaction vector of $k$-th order over testing samples $\overline{\mathbf{I}}^{(k)}_{c, \text{test}}$ to evaluate the generalization power of interactions of $k$-th order \emph{w.r.t.} the classification of the category $c$, \emph{i.e., }
$\text{Sim}(\overline{\mathbf{I}}^{(k)}_{c, \text{train}}, \overline{\mathbf{I}}^{(k)}_{c, \text{test}}) = \frac
{\| \min(\widehat{\mathbf{I}}^{(k)}_{c, \text{train}}, \widehat{\mathbf{I}}^{(k)}_{c, \text{test}}  ) \|_1} 
{\| \max(\widehat{\mathbf{I}}^{(k)}_{c, \text{train}}, \widehat{\mathbf{I}}^{(k)}_{c, \text{test}}) \|_1}$, where $\widehat{\mathbf{I}}^{(k)}_{c, \text{train}} = [(\max(\overline{\mathbf{I}}^{(k)}_{c, \text{train}}), \mathbf{0})^T, (\max(-\overline{\mathbf{I}}^{(k)}_{c, \text{train}}), \mathbf{0})^T]^T$ and $\widehat{\mathbf{I}}^{(k)}_{c, \text{test}} = [(\max(\overline{\mathbf{I}}^{(k)}_{c, \text{test}}), \mathbf{0})^T, (\max(-\overline{\mathbf{I}}^{(k)}_{c, \text{test}}), \mathbf{0})^T]^T$ project two $2^d$-dimensional interaction vector to two $2^{d+1}$-dimensional non-negative vectors to enable the computation of the Jaccard similarity. A large Jaccard similarity indicates a stronger generalization power.

\begin{figure}[t]
    \centering
    \includegraphics[width=0.99\textwidth, height=6cm]{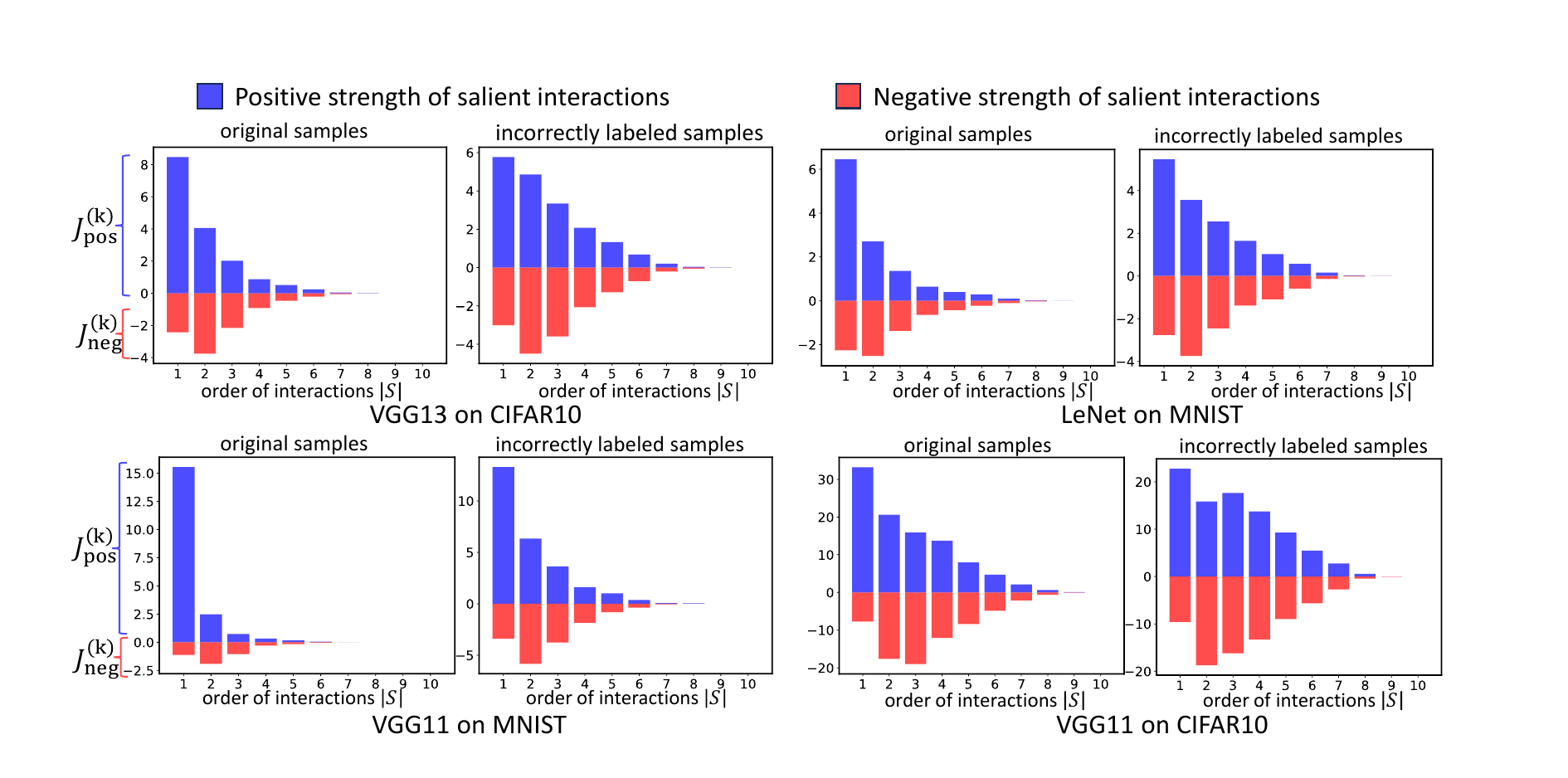}
    \caption{The distribution of interactions over different orders extracted from $100$ original samples and extracted from $100$ incorrectly labeled samples.}
    \label{img-high-order}
\end{figure}

We conducted experiments to compute $\text{Sim}(\overline{\mathbf{I}}^{(k)}_{c, \text{train}}, \overline{\mathbf{I}}^{(k)}_{c, \text{test}})$ 
for interactions of different orders. We tested LeNet trained on the MNIST dataset, VGG-11 trained on the CIFAR-10 dataset, VGG-13 trained on the CUB200-2011 dataset and VGG-13 trained on the Tiny-ImageNet dataset. In order to reduce the computational cost, we only computed the mean of the Jaccard similarity $\mathbb{E}_{c}[\text{Sim}(\overline{\mathbf{I}}^{(k)}_{c, \text{train}}, \overline{\mathbf{I}}^{(k)}_{c, \text{test}})]$ over the first 10 categories. Figure \ref{img-similarity} shows that the Jaccard similarity of the interactions kept decreasing as the order of the interactions increased. Thus we verify that the generalization power of high-order interactions was usually weaker than that of low-order interactions.

\textbf{\textbullet \  Further evidence that high-order interactions have weak generalization power.} We visualized the distribution of interactions over different orders encoded by a DNN, which was trained with noise labels. Specifically, we trained VGG-11 on the MNIST dataset, VGG-13 on the MNIST dataset, and VGG-13 on the CIFAR-10 dataset. In particular, we assigned some ($0.17\%$) training samples with an incorrect label. In this way, the original samples in the dataset could be considered simple samples, while a few samples with incorrect labels corresponded to hard samples that were supposed to cause over-fitting of the DNN. The details of how to assign training samples with an incorrect label are provided in Appendix \ref{sec:noise-label}.

Figure \ref{img-high-order} compares the interaction's complexity (orders) between interactions extracted from original samples and interactions extracted from incorrectly labeled samples. We found that LeNet, VGG-11, and VGG-13 used much more complex interactions (interactions of much higher orders) for the classification of incorrectly labeled samples than the classification of original samples. This further validated that high-order interactions usually had weaker generalization power than low-order interactions.

\textbf{Discussions: Interactions can explain the underlying metrics for the generalization power of a DNN.} (1) The first metric is the order (complexity) of interactions. \citet{zhou2024explaining} have found that high-order interactions have poor generalization power. If a DNN encodes too many high-order interactions for a given input sample, it suggests that the DNN has poor generalization power on this sample. (2) For interactions of each order, a positive and negative offset of these interactions is another metric that indicates the generalization power of a DNN. If interactions extracted from a given input sample almost offset each other, then the DNN may have poor generalization power on this sample. These two metrics provide new insights into the generalization power of the DNN. Unlike traditional studies~\cite{dziugaite2017computing, neyshabur2015norm, keskar2016large, foret2020sharpness, kwon2021asam}, the order and the positive and negative offset of interactions first bridge the generalization power of an entire black-box model with its detailed inference patterns (interactions).

\textbf{Our research shows that a high confidence score does not always represent a faithful inference.} Let us consider the following two cases. (1) The DNN's inference with a high classification confidence may not necessarily be reliable, because the DNN may use considerable high-order interactions for inference, or has a significant offset between positive interactions and negative interactions. Then, this DNN is probably over-fitted.
(2)Not-so-confident classifications may either not be unreliable. Sometimes, the DNN may have learned very few interactions to classify an input sample, but all learned interactions are simple and generalizable.

\subsection{Proof and discussion}
\textbf{Proof the fusiform-like distribution of interactions over different orders when network parameters are randomly initialized.} We consider that the initialized DNN mainly encodes noisy patterns, thereby generating random interactions. Thus, Theorem \ref{throrem:3} proves that the distribution of interactions encoded by the initialized DNN has a fusiform shape.

\begin{theorem}[proved in \ref{prove:throrem:3}]\label{throrem:3}
    Let us assume that all interactions encoded by a DNN with randomly initialized parameters represent noises, and follow a Gaussian distribution $\forall S\subseteq N, I(S\vert \mathbf{x}) \sim \mathbb{N}(\mathbf{0}, \sigma^2 \cdot \mathbf{I})
    $, where $\mathbf{I}$ represents identity matrix. Then, let $\Psi_{\text{pos}}^{(k)} = \sum\nolimits_{S \subseteq N:\vert S \vert = k} \max(I(S \vert \mathbf{x}),0)$ and $\Psi_{\text{neg}}^{(k)} = \sum\nolimits_{S \subseteq N:\vert S \vert = k} \min(I(S \vert \mathbf{x}),0)$ denote the strength of all positive and the strength of all negative AND-OR interactions of the $k$-th order. Then, the mean of the $\Psi_{\text{pos}}^{(k)}$ and $\Psi_{\text{neg}}^{(k)}$ encoded by an initialized DNN are
    \begin{align}
        \mathbb{E}[\Psi_{\text{pos}}^{(k)}] =  \tbinom{n}{k} \cdot \sqrt{\sigma / 2\pi}, \ \ \ \ \  \mathbb{E}[\Psi_{\text{neg}}^{(k)}] =  -\tbinom{n}{k} \cdot \sqrt{\sigma / 2\pi}
    \end{align}
\end{theorem}

\textbf{Alignment between the two-phase phenomenon and the gap between testing and training losses during the training process.} The gap between the training and testing losses is the most widely used metric for the over-fitting level of a model. To this end, we have found that the two-phase phenomenon and the gap between the testing and training losses are aligned temporally during the training process. We followed the experimental settings in Section \ref{sec:experiments}, and we found that the two-phase phenomenon and the gap between the testing and training losses are aligned temporally during the training process. 
Figure \ref{img-two-phase-experiment} shows the curves of the training loss, the testing loss, and the gap between the two losses. It also shows the distribution of interactions over different epochs. 
We also annotated the epoch when the gap between the testing and training losses began to increase The annotated epoch was taken as the end of the first phase and the beginning of the second phase.
All these DNNs exhibited the two-phase phenomenon, which aligned the gap between the testing and training losses.

In this way, we can understand the above phenomenon as follows. Before the training process, the interactions encoded by the initialized DNN all represented random noises, and the distribution of the interactions over different orders looked like a fusiform. During the first phase, the DNN penalized interactions of medium and high orders, and gradually learned the simplest (low-order) interactions. In particular, most DNNs removed high-order interactions and only encoded interactions of the lowest orders just before the annotated epoch (\emph{i.e.,} the epoch before the gap between the testing and training losses). Then, in the second phase, the DNN encoded interactions of gradually increasing orders. Because \citet{zhou2024explaining} have found that high-order interactions usually had weaker generalization power than low-order interactions, we could consider in the second phase, the DNN first learned interactions with the strongest generalization power and then gradually shifted its attention to a bit more complex yet less generalizable interactions. Some DNNs were finally over-fitted and encoded many interactions of medium and high orders.

\textbf{Identifying the starting point of learning over-fitted features.}
For most shallow models (\emph{e.g.} the support vector machine), the generalization power of a model is usually considered an intrinsic property of the model over the entire testing dataset. However, for deep models, the generalization power of a DNN has become a more complex problem. For example, \cite{zhou2024explaining} have found that given different input samples, a DNN may exhibit fully different generalization powers. According to both the findings in \cite{zhou2024explaining} and experiments in Figure \ref{img-similarity} and \ref{img-high-order}, given some input samples, the DNN uses low-order interactions for inference, and we can consider that the inference is conducted on relatively simple and generalizable features. Whereas, given other input samples, the DNN triggers interactions of medium and high orders, and we can consider that the DNN uses over-fitted features for inference.

In this way, interactions offer a new perspective to analyze the DNN's generalization power on each input sample making the under-fitting and over-fitting no longer two contradictory issues.
\textit{Emprically, a DNN usually begins to learn over-fitted features shortly after entering the second phase of the training process.} 

\textbullet \ Under-fitting can be understood as the lack of sufficient generalizable interactions (usually low-order interactions), while over-fitting is referred to as the encoding of non-generalizable interactions (usually high-order interactions).

\textbullet \ The beginning of learning over-fitted features does not mean the stop of learning meaningful features. Over-fitted features and normal features may be simultaneously learned in the training process. \emph{I.e.,} low-order interactions and medium/high-order interactions are optimized simultaneously in the second phase.

\section{Conclusion}
In this paper, we have used interactions to explain the primitive inference patterns used by the DNN, and we have discovered the two-phase dynamics of a DNN learning interactions of different complexities (orders). Specifically, we have discovered and later proven that before the training process, a DNN with randomly initialized parameters mainly encodes interactions of medium orders. Then, the training process has two phases. The first phase mainly penalizes interactions of medium and high orders, and the second phase mainly learns interactions of gradually increasing orders. More interestingly, the two-phase dynamics is temporally aligned with the change of the gap between the testing and training losses during the learning process. In other words, the two-phase dynamics of interactions can be considered as a fine-grained mechanism for the change of the generalization power of a DNN during the training process. In other words, the discovered two-phase dynamics illustrates how a DNN learns detailed generalizable and over-fitted interactions in different epochs, and how a DNN's feature representation changes from under-fitting to over-fitting. Various ablation studies have proven that high-order interactions usually have weaker generalization power than low-order interactions. Then, we have conducted experiments to extract interactions from DNNs with various architectures trained for different tasks. The two-phase dynamics have been successfully verified on all these DNNs.

\bibliographystyle{plainnat}
\bibliography{references}

\newpage
\appendix

\section{Experimental detail}
\subsection{Training settings.} In this paper, we trained various DNNs for different tasks. Specifically, for the image classification task, we trained LeNet and VGG-11/13 on the MNIST dataset with a learning rate of 0.01. We trained LeNet, and VGG-11/13 on the CIFAR-10 dataset with a learning rate of 0.01. We trained AlexNet, VGG-13/16 on the CUB200-2011 dataset with a learning rate of 0.01. We trained AlexNet, VGG-13/16 on the Tiny-ImageNet dataset with a learning rate of 0.001. For natural language processing tasks, we trained the Bert-Tiny model and Bert-Medium model on the SST-2 dataset with a learning rate of 0.01. For point-cloud classification tasks, we trained DGCNN on the ShapeNet dataset with a learning rate of 0.01. All DNNs were trained using the SGD optimizer and were trained for 512 epochs.

\subsection{Details about how to calculate interactions for different DNNs.}
\textbf{\textbullet \ For image data in different image datasets,} since the computational cost of interactions was intolerable, we applied a sampling-based approximation method to calculate $I(S \vert \mathbf{x})$. Specifically, we considered the output after the second ReLU layer for VGG-11, VGG-13, and VGG-16 and masked the output after the first ReLU layer for the other DNNs as middle features of DNNs. For the CIFAR-10 dataset, the CUB200-2011 dataset, and the Tiny-ImageNet dataset, we uniformly split each middle feature into $8 \times 8$ patches. Furthermore, we randomly sampled 10 patches from the central $6 \times 6$ region to calculate interactions (\emph{i.e.,} we did not sample patches that were on the edges of an image), and considered these patches as input variables for each middle feature. Similarly, for the MNIST dataset, we uniformly split each input middle feature into $7 \times 7$ patches and randomly sampled 10 patches from the central $5 \times 5$ region. We used $\mathbf{0}$ as a baseline value to mask the variables in $N \backslash T$.

\textbf{\textbullet \ For natural language processing data in SST-2 dataset,} we considered the input tokens as input variables for each input sentence, and we randomly sampled 10 tokens to calculate interactions. We used the ``mask'' token with the token id $= 103$ to mask the tokens in $N \backslash T$. 

\textbf{\textbullet \ For 3D point cloud data in ShapeNet datasets,} we clustered all the points into 30 clusters, and considered these clusters as input variables for each 3D point cloud. we finally randomly selected 10 clusters as variables to calculate interactions. We use the average value of each cluster to mask the corresponding cluster in $N \backslash T$.

\section{Detail of three conditions for the relatively stable output of a DNN.}\label{sec:condition}
\citet{ren2023we} have formulated the sparsity property of AND interactions, and there are three conditions in mathematics as follows.

\textbf{Condition 1.} \textit{Interactions beyond the $M$-th order were not encoded by the DNN: {\small$\forall \ S\in \{S\subseteq N \mid \vert S\vert \ge M+1\}, \ I_{\text{and}}(S|\mathbf{x})=0$}.}

Condition 1 suggests that interactions beyond the $M$-th order were not encoded by the DNN.
This is because such interactions typically denote very complex and over-fitted patterns, which are unnecessary and unlikely for the DNN to learn in practical applications.

\textbf{Condition 2.} \textit{Let us consider the average network output over all masked samples $\mathbf{x}_S$ with $|S|=k$ unmasked input variables. This average network output monotonically increases when $k$ increases: $\forall \ k' \le k$, we have {\small$\bar{u}^{(k')} \le \bar{u}^{(k)}$}, where {\small$\bar{u}^{(k)}\overset{\text{\rm def}}{=}\mathbb{E}_{|S|=k}[v(\mathbf{x}_S)-v(\mathbf{x}_\emptyset)]$}.}

Condition 2 implies that a well-trained DNN is likely to have higher classification confidence for input samples that are less masked.

\textbf{Condition 3.} \textit{Given the average network output $\bar{u}^{(k)}$ of samples with $k$ unmasked input variables, there is a polynomial lower bound for the average network output of samples with $k' (k'\le k)$ unmasked input variables: {\small $\forall \ k' \le k, \ \bar{u}^{(k')} \ge (\frac{k'}{k})^p \ \bar{u}^{(k)}$}, where $p>0$ is a positive constant.}

Condition 3 implies that the classification confidence of the DNN does not significantly degrade
on masked input samples. The classification/detection of masked/occluded samples is common in real scenarios. In this way, a well-trained DNN usually learns to classify a masked input sample based on local information (which can be extracted from unmasked parts of the input) and thus should not yield a significantly low confidence score on masked samples.

\section{Strategies of adding noise labels}\label{sec:noise-label}
We trained VGG-11 and LeNet on the MNIST dataset, then we used the well-trained VGG-11 and LeNet to find 100 samples with the lowest classification confidence for each category in the training set separately, and then set their corresponding labels to the second-best category.

Similarity, We trained VGG-11 and VGG-13 on the CIFAR-10 dataset, then we used the well-trained VGG-11 and VGG-13 to find 100 samples with the lowest classification confidence for each category in the training set separately, and then set their corresponding labels to the second-best category.

\section{Proof details}
\subsection{Proof of Theorem 3 in the main paper}\label{prove:throrem:3}
\textbf{Theorem 3.} Let us assume that all interactions encoded by a DNN with randomly initialized parameters represent noises, and follow a Gaussian distribution $\forall S\subseteq N, I(S\vert \mathbf{x}) \sim \mathbb{N}(\mathbf{0}, \sigma^2 \cdot \mathbf{I})$, where $\mathbf{I}$ represents identity matrix. Then, let $\Psi_{\text{pos}}^{(k)} = \sum\nolimits_{S \subseteq N:\vert S \vert = k} \max(I(S \vert \mathbf{x}),0)$ and $\Psi_{\text{neg}}^{(k)} = \sum\nolimits_{S \subseteq N:\vert S \vert = k} \min(I(S \vert \mathbf{x}),0)$ denote the strength of all positive and the strength of all negative AND-OR interactions of the $k$-th order. Then, the mean of the $\Psi_{\text{pos}}^{(k)}$ and $\Psi_{\text{neg}}^{(k)}$ encoded by an initialized DNN are
\begin{align*}
    \mathbb{E}[\Psi_{\text{pos}}^{(k)}] =  \tbinom{n}{k} \cdot \sqrt{\sigma / 2\pi}, \ \ \ \ \  \mathbb{E}[\Psi_{\text{neg}}^{(k)}] =  -\tbinom{n}{k} \cdot \sqrt{\sigma / 2\pi}
\end{align*}
\begin{proof}
    Let $d = \tbinom{n}{k}$ denotes the number of interactions of $k$-th order.
    Let us assume that all interactions encoded by a DNN with randomly initialized parameters represent noises, and follow a Gaussian distribution $\forall S\subseteq N, I(S\vert \mathbf{x}) \sim \mathbb{N}(\mathbf{0}, \sigma^2 \cdot \mathbf{I})$, where $\mathbf{I}$ represents identity matrix. Then, the mean absolute deviation (MAD) of all interactions $\mathbb{E}[\vert I(S\vert \mathbf{x}) \vert]$ is equal to $\sqrt{2 \sigma / \pi}$.
    In this way,
    \begin{align*}
        \mathbb{E}[\Psi_{\text{pos}}^{(k)}] &= \sum\nolimits_{S \subseteq N:\vert S \vert = k} \max(I(S \vert \mathbf{x}),0) \\ &=\tbinom{n}{k} \cdot P(I(S\vert \mathbf{x}) > 0) \cdot \mathbb{E}[\vert I(S\vert \mathbf{x}) \vert] \\
        &= \tbinom{n}{k} \cdot \frac{1}{2} \cdot \sqrt{2 \sigma / \pi} \\
        &= \tbinom{n}{k} \cdot \sqrt{\sigma / 2\pi},\\
        \mathbb{E}[\Psi_{\text{neg}}^{(k)}] &= \sum\nolimits_{S \subseteq N:\vert S \vert = k} \min(I(S \vert \mathbf{x}),0) \\ &=\tbinom{n}{k} \cdot P(I(S\vert \mathbf{x}) < 0) \cdot -\mathbb{E}[\vert I(S\vert \mathbf{x}) \vert] \\
        &= \tbinom{n}{k} \cdot \frac{1}{2} \cdot -\sqrt{2 \sigma / \pi} \\
        &= -\tbinom{n}{k} \cdot \sqrt{\sigma / 2\pi}.
    \end{align*}
    Therefore, Theorem 3 holds.
\end{proof}

\subsection{Proof that the AND-OR component only contains AND-OR interactions.}\label{sec:match}
The DNN's output $v(\mathbf{x}_T)$ can be decomposed into two components $v(\mathbf{x}_T) = v_{\text{and}}(\mathbf{x}_T) + v_{\text{or}}(\mathbf{x}_T) + v(\emptyset)$. It is easy to prove that the component $v_{\text{and}}(\mathbf{x}_T)$ only contains AND interactions $v_{\text{and}}(\mathbf{x}_T) = \sum_{\emptyset \ne S^{\prime} \subseteq T} I_{\text{and}}(S^{\prime}\vert \mathbf{x}_T)$, and the component $v_{\text{or}}(\mathbf{x}_T)$ only contains OR interactions $v_{\text{or}}(\mathbf{x}_T) = \sum_{S\subseteq N:S \cap T \neq \emptyset} I_{\text{or}}(S\vert \mathbf{x}_T)$.
\begin{proof}
    According to the definition of AND interaction, we have $\forall T \subseteq N$,
    \begin{small}
    \begin{align*}
        \sum_{S^{\prime} \subseteq T}I_{\text{and}}(S^{\prime}\vert \mathbf{x}) &= \sum_{S^{\prime}\subseteq T} \sum_{L \subseteq S^{\prime}} (-1)^{\vert S^{\prime} \vert - \vert L \vert} (v_{\text{and}}(L) - v_{\text{and}}(\emptyset)) \\
        &= \sum_{L \subseteq T} \sum_{S^{\prime}\subseteq T : S^{\prime} \supseteq L}(-1)^{\vert S^{\prime} \vert -\vert L\vert} (v_{\text{and}}(L) - v_{\text{and}}(\emptyset)) \\
        &= \sum_{L \subseteq T} \sum_{t=\vert L\vert}^{\vert T\vert}\sum_{S^{\prime}\subseteq T: T\supseteq L, \vert S^{\prime}\vert =t}(-1)^{t -\vert L\vert} (v_{\text{and}}(L) - v_{\text{and}}(\emptyset)) \\
        &= \sum_{L \subseteq T} (v_{\text{and}}(L) - v_{\text{and}}(\emptyset)) \sum_{k=0}^{\vert T\vert - \vert L\vert} \binom{\vert T\vert - \vert L\vert}{k}(-1)^k \\
        &= v_{\text{and}}(\mathbf{x}_T) - v_{\text{and}}(\emptyset).
    \end{align*}
    \end{small}
    Therefore, we have $v_{\text{and}}(\mathbf{x}_T) = \sum_{S^{\prime} \subseteq T} I_{\text{and}}(S^{\prime}\vert \mathbf{x})$. According to the definition of OR interaction, we have $\forall T \subseteq N$,
    \begin{small}
    \begin{align*}
        \sum_{S\subseteq N:S \cap T \neq \emptyset} I_{\text{or}}(S\vert \mathbf{x}_T)
        &= \sum_{S\subseteq N:S \cap T \neq \emptyset} \left[- \sum\nolimits_{L \subseteq S} (-1)^{\vert S \vert - \vert L \vert} v_{\text{or}}(\mathbf{x}_{N \setminus L}) \right]\\
        &= - \sum\nolimits_{L \subseteq N} \sum\nolimits_{S: S \cap T \neq \emptyset, S \supseteq L} (-1)^{\vert S \vert - \vert L \vert} v_{\text{or}}(\mathbf{x}_{N \setminus L}) \\
        &=  - \left[\sum_{\vert S' \vert = 1}^{\vert T \vert} C_{\vert T \vert}^{\vert S' \vert} (-1)^{\vert S' \vert} \right] \cdot \underbrace{v_{\text{or}}(\mathbf{x}_T)}_{L=N\setminus T} - \underbrace{v_{\text{or}}(\mathbf{x}_{\emptyset})}_{L=N} \\
        &\quad- \sum_{L \cap T \neq \emptyset, L \neq N} \left[\sum_{S' \subseteq N\setminus T \setminus L} \left( \sum_{\vert S'' \vert = 0}^{\vert T \vert-\vert T \cap L \vert} C_{\vert T \vert - \vert T \cap L \vert}^{\vert S''\vert } (-1)^{\vert S' \vert + \vert S'' \vert} \right) \right]\cdot v_{\text{or}}(\mathbf{x}_{N \setminus L})  \\
        &\quad- \sum_{L \cap T=\emptyset, L \neq N \setminus T} \left[ \sum_{S' \subseteq N\setminus T \setminus L} \left( \sum_{\vert S'' \vert=0}^{\vert T \vert} C_{\vert T \vert}^{\vert S'' \vert} (-1)^{\vert S' \vert + \vert S'' \vert}\right) \right] \cdot v_{\text{or}}(\mathbf{x}_{N \setminus L})  \\
        &=  - (-1) \cdot v_{\text{or}}(\mathbf{x}_T) - v_{\text{or}}(\mathbf{x}_{\emptyset}) - \sum_{L \cap T \neq \emptyset, L \neq N} \left[\sum_{S' \subseteq N\setminus T \setminus L} 0 \right]\cdot v_{\text{or}}(\mathbf{x}_{N \setminus L})  \\
        &\quad- \sum_{L \cap T=\emptyset, L \neq N \setminus T}\left[\sum_{S' \subseteq N\setminus T \setminus L} 0 \right] \cdot v_{\text{or}}(\mathbf{x}_{N \setminus L})  \\
        &= v_{\text{or}}(\mathbf{x}_T) - v_{\text{or}}(\mathbf{x}_{\emptyset})
    \end{align*}
    \end{small}
Therefore, we have $v_{\text{or}}(\mathbf{x}_T) = \sum_{S\subseteq N:S \cap T \neq \emptyset} I_{\text{or}}(S\vert \mathbf{x}_T)$.
Then, the component $v_{\text{and}}(\mathbf{x}_T)$ only contains AND interactions $v_{\text{and}}(\mathbf{x}_T) = \sum_{S^{\prime} \subseteq T} I_{\text{and}}(S^{\prime}\vert \mathbf{x})$, and the component $v_{\text{or}}(\mathbf{x}_T)$ only contains OR interactions $v_{\text{or}}(\mathbf{x}_T) = \sum_{S\subseteq N:S \cap T \neq \emptyset} I_{\text{or}}(S\vert \mathbf{x}_T)$.
\end{proof}

\end{document}